# Efficient Behavior-aware Control of Automated Vehicles at Crosswalks using Minimal Information Pedestrian Prediction Model

Suresh Kumaar Jayaraman[1], Lionel P. Robert Jr.[2], Xi Jessie Yang[3], Anuj K. Pradhan[4], and Dawn M. Tilbury[1]

*Abstract*— For automated vehicles (AVs) to reliably navigate through crosswalks, they need to understand pedestrians' crossing behaviors. Simple and reliable pedestrian behavior models aid in real-time AV control by allowing the AVs to predict future pedestrian behaviors. In this paper, we present a Behavior-aware Model Predictive Controller (B-MPC) for AVs that incorporates long-term predictions of pedestrian crossing behavior using a previously developed pedestrian crossing model. The model incorporates pedestrians' gap acceptance behavior and utilizes minimal pedestrian information, namely their position and speed, to predict pedestrians' crossing behaviors. The B-MPC controller is validated through simulations and compared to a rule-based controller. By incorporating predictions of pedestrian behavior, the B-MPC controller is able to efficiently plan for longer horizons and handle a wider range of pedestrian interaction scenarios than the rule-based controller. Results demonstrate the applicability of the controller for safe and efficient navigation at crossing scenarios.

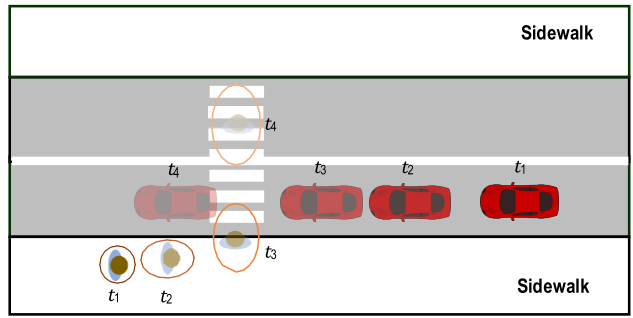

Fig. 1. Representation of a typical interaction between a vehicle and a crossing pedestrian. The vehicle has to plan its trajectory considering the moving pedestrian and by following the center line of the road. The illustration shows the predicted pedestrian trajectory and the uncertainty ellipses at various time instances and the planned trajectory of the vehicle.

## I. INTRODUCTION

An important challenge for automated vehicles (AVs) is driving in urban environments because AVs have to interact with pedestrians to avoid collisions and facilitate smooth traffic flow and pedestrian movements can change instantaneously [1]. Interactions with pedestrians at unsignalized crosswalks is particularly challenging as the right-of-way is unclear, making it hard to predict pedestrians' actions. To ensure safety, AVs are expected to drive cautiously around pedestrians [2], which, however, can encourage pedestrians to have careless beahviors like stepping onto the road to force the risk-averse AVs to slow down and yield.

Behavior-aware control, which anticipates the behaviors of pedestrians, can help AVs plan actions that are safe, yet less conservative [3], [4]. Fig. 1 illustrates the AV behavior-aware control problem for a typical interaction with a pedestrian at an unsignalized mid-block crosswalk. Studies have incorporated pedestrian behavior into AV control by assuming pedestrians as moving obstacles with a constant velocity and Gaussian noise [5], [6]. This simple approach is effective for short duration planning ($\leq 2$ s) and when the pedestrians are moving. However, at crosswalks, pedestrian behavior is much more unpredictable as they have to wait for an opportunity and decide when to cross.

There are two primary challenges in pedestrian behavior aware AV control: (i) developing simple pedestrian prediction models that run in real-time while still capturing the interactions between pedestrians and AVs, and (ii) effectively incorporating these predictions in an AV control framework. In this paper, we address these challenges by including predictions of pedestrian crossing behavior from a pedestrian crossing model previously developed in [7]. We model pedestrians as a hybrid automaton that switches between discrete actions. We develop a Behavior-aware Model Predictive Controller (B-MPC) that utilizes predictions from the pedestrian crossing model for optimal AV control.

The main contribution of this paper is developing a behavior-aware controller for real-time AV motion planning in urban environments that uses a simple pedestrian crossing model. The controller was able to avoid collisions with pedestrians, plan for a long duration ($\leq 5$ s), and handle a wider range of scenarios (wider range of pedestrians crossing gaps without collisions), compared to the baseline controller.

The rest of this paper is organized as follows. Section II explains the background and existing work on behavior-aware AV control. Section III explains our optimization and control framework and Section IV details the simulation setup. Section V explains the results followed by Conclusion and Future work in Section VI.

[1]Suresh Kumaar Jayaraman (*corresponding author*) and Dawn M. Tilbury are with the Department of Mechanical Engineering, University of Michigan, Ann Arbor, MI 48109, USA (e-mail: jskumaar@umich.edu; tilbury@umich.edu)

[2]Lionel P. Robert Jr. is with the School of Information, University of Michigan, Ann Arbor, MI 48109, USA (e-mail: lprobert@umich.edu)

[3]Xi Jessie Yang is with the Department of Industrial and Operations Engineering, University of Michigan, Ann Arbor, MI 48109, USA (e-mail: xijyang@umich.edu)

[4]Anuj K. Pradhan is with the Department of Mechanical and Industrial Engineering, University of Massachusetts, Amherst, MA 01003, USA (e-mail: anujkpradhan@umass.edu)

## II. RELATED WORK AND BACKGROUND

Behavior-aware control allows AVs to anticipate behaviors of other traffic participants and plan accordingly. For example, [8] implemented a POMDP-based planner on an autonomous electric cart. The planner was able to avoid collisions with pedestrians in unstructured environments (university campus) by utilizing a social force model for the pedestrians. Similarly, [9] used an LQR-based pedestrian model to plan AV motion using a Model Predictive Controller (MPC). These models demonstrate the efficacy of incorporating pedestrian predictions into AV control. However, neither approaches explicitly included the decision-making process of the pedestrian to cross the street and their corresponding stop-and-go behavior, limiting their application at crosswalk interactions.

Pedestrian models that predict crossing decisions have been developed previously [10]–[12]. These approaches utilize pedestrian's pose, motion, and vehicle behavior to develop Markovian [11] or Neural Network models [10], [12]. However, these models require a large amount of data with rich information, such as the pedestrian's pose. Recently, reinforcement learning approaches have been used for AV control that incorporate the interactive behavior between AVs and other road users [3], [13]. The AVs were automatically able to calculate actions that aid negotiation between the AV and road users at intersections to avoid collisions between them. However, such methods require extensive amounts of data. Model predictive controllers (MPC), on the other hand, have been proved to be useful for generating collision-free AV trajectories with less data requirements [14]–[17]. For example, Werling and Liccardo [17] determined evasive trajectories using nonlinear model-predictive control (NMPC). However, the above study assumed future actions of the pedestrian to be completely known. In this paper, we do not make this assumption, but forecast the position of the pedestrian using a pedestrian model.

## III. PROBLEM FORMULATION

Consider the scenario where an AV is approaching an unsignalized crosswalk as shown in Fig. 1. Pedestrians approaching the crosswalk decide to either cross the road or wait for the AV to pass. The AV has to plan actions that not only ensure safety but also help the riders in the AV reach their destinations comfortably. Using the pedestrian crossing model developed in [7], AVs can predict future pedestrian states and plan their actions accordingly. In the following, we explain the AV and pedestrian models used and formulate a receding horizon model predictive control problem to calculate the AV control inputs.

### A. AV Model

Dynamic vehicle models are comprehensive but challenging to use for real-time AV motion planning, especially in urban scenarios. In addition to being computationally expensive, the tire models have vehicle velocity in the denominator for tire slip estimation and become singular at low speeds. Hence, these models are not suitable for stop-and-go scenarios common in urban driving [18]. Thus, we assume the AV to be a point mass with a rectangular footprint (refer Fig. 4). We assume that longitudinal vehicle dynamics is sufficient for the crosswalk interactions and employed a discrete-time kinematic model shown in equation (1), where $\mathbf{X} = [x_v \ v_v]^T$ is the state vector comprising the position and velocity of the vehicle respectively. $\Delta t$ denotes the discretization time step and $a_v$ is the acceleration input that governs the AV's motion.

$$x_{v_{k+1}} = x_{v_k} + \Delta t \ v_{v_k} \quad (1)$$
$$v_{v_{k+1}} = v_{v_k} + \Delta t \ a_{v_k}$$

### B. Pedestrian Crossing Model

We developed a simple pedestrian model applicable to crosswalk scenarios. A main advantage of this model, when compared to existing pedestrian crossing models [9], [19] is its ability to predict pedestrian crosswalk behavior for long durations. Further, this model uses minimum information, namely the pedestrian's position and velocity, and does not require information about pedestrian actions or pose. The model development procedure is detailed in [7], but for the sake of completeness, we briefly describe the model below.

We modeled pedestrian crossing behavior (refer Fig. 2) as a probabilistic hybrid automaton. At any instant, the pedestrian can be in one of the four discrete states (actions) – *Approach Crosswalk*($q_1$), *Wait*($q_2$), *Cross*($q_3$), or *Walk away*($q_4$) (from crosswalk). Pedestrians need to make a decision to cross or wait whenever they are approaching the crosswalk or are already waiting at the crosswalk. We express the probabilistic transitions between the *Approach Crosswalk* and *Wait/Cross* states, i.e. $p(q_2|q_1)$ or $p(q_3|q_1)$ and the probabilistic transition between *Wait* and *Cross* state, i.e. $p(q_3|q_2)$ using pedestrian's gap acceptance behavior [20]. Pedestrians evaluate the available time gap to cross the street and either accept the gap by starting to cross or reject the gap by waiting at the crosswalk [20]. Pedestrian motion in each of the four discrete states was expressed using a constant velocity point mass model with Gaussian noise.

The gap acceptance model allowed us to express the transitions $p(q_2|q_1)$, $p(q_3|q_1)$, and $p(q_3|q_2)$ as a single decision-making process, i.e. *what is the probability of accepting the current traffic gap?* We defined gap as the time taken by the vehicle, traveling at its current velocity, to reach the pedestrian's longitudinal position along the road. We considered the following assumptions in developing the gap acceptance model.

*Assumption 1:* A gap is accepted when the pedestrian starts walking laterally to cross the street during that gap.

*Assumption 2:* Pedestrians crossing the street exhibit rational behavior and always use the crosswalk. Thus, gaps are only accepted when the pedestrian is close to the crosswalk, defined by the decision zone in Fig. 3.

*Assumption 3:* The decision to accept/reject a gap is always made at the start of a gap, and the decision holds for the entire duration of that gap. A gap can only start when the previous vehicle has just passed the pedestrian (refer Fig. 3). Thus, when pedestrians reach the decision zone during the middle of a gap, they wait for the next gap near the crosswalk to make the decision.

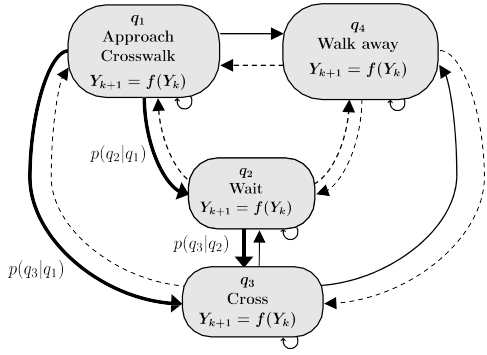

Fig. 2. Hybrid automaton model of pedestrian crossing behavior. $q_i$ denotes the discrete actions, $Y_k$ the continuous pedestrian states at time step $k$, and $p(q_i|q_j)$ the probability of transition between the states. The solid arrows (bold and unbold) indicate common transitions for a rational pedestrian with or without the intention to cross. The bold arrows indicate the transitions corresponding to pedestrian gap acceptance behavior. The dotted arrows indicate all other possible pedestrian behaviors.

TABLE I
INPUT PARAMETERS FOR GAP ACCEPTANCE MODEL.

| Parameter | Description |
| --- | --- |
| Gap Duration [s] | Expected time taken by vehicle to reach pedestrian at current instantaneous speed ($d_m/v_v$) |
| Wait Time [s] | Time elapsed since pedestrian started waiting |
| Curb Distance [m] | Lateral distance between pedestrian and road edge |
| Crosswalk Distance [m] | Longitudinal distance between pedestrian and center of crosswalk |
| Speed [m/s] | Average pedestrian speed in the previous second |

We modeled the gap acceptance behavior using a Support Vector Machine (SVM) classifier and obtained probabilistic outputs from the SVM model following the method in [21]. The model used the inputs detailed in Table I. These parameters significantly affect pedestrian crossing behavior [20], [22] and are easy to process in real-time. The gap acceptance model was combined with a constant velocity dynamics and a Kalman filter framework to predict pedestrian trajectory including the prediction uncertainty, for future time steps. The data used for developing the model was collected from pedestrian interactions with AVs in a virtual reality environment [7], [23]. More information on the model can be found in [7] for interested readers.

### C. Problem Formulation

The Behavior-aware Model Predictive Controller (B-MPC) calculates the inputs to achieve the AVs' objectives expressed through a cost function. The physical limitations of the AV and collision avoidance with pedestrians are expressed as constraints. The problem is formulated as a constrained quadratic optimization problem, which enables fast computation of control inputs, suitable for real-time planning. The optimization problem is formalized in equation (2), where $J$ is the cost function and $\mathbf{Z} = [\mathbf{X}\ \mathbf{V}\ \mathbf{U}\ \Delta\mathbf{U}\ \mathbf{R}]^T$ is a stacked vector of all states, control inputs, change in control, and references, respectively, for horizon $N$.

$$\begin{aligned} \min_{\mathbf{Z}} \quad & J(\mathbf{Z}) \\ s.t. \quad & A_{eq}\mathbf{Z} = B_{eq} \\ & A_{ineq}\mathbf{Z} \leq B_{ineq} \\ & l_b \leq \mathbf{Z} \leq u_b \end{aligned} \quad (2)$$

### D. Cost Function

Safety is the main priority in the AV control problem. However, the AVs should also follow speed limits and reach their destination on time while maintaining ride comfort. The quadratic cost matrices $Q$ for the various objectives are constructed using their corresponding weights $w$, which are chosen to be positive to ensure the matrices $Q$ are positive semi-definite. The objective cost function is given by

$$J = J_{target} + J_{jerk} + J_{acc} + J_{speed}. \quad (3)$$

*1) Target Cost:* One of the primary objectives of the AV is to reach a target destination. Since pedestrians always cross the street at the crosswalk (refer Assumption 2), for the purpose of simplicity, we consider the destination to be an arbitrary point $x_v^{ref}$ beyond the crosswalk (refer Fig. 5). This ensures that passing the crosswalk is one of the objectives of the AV. The difference between the destination $x^{ref}$ and the vehicle position at the end of the prediction horizon $x^N$ is penalized as $J_{target} = (x_v^N - x_v^{ref})^T Q_{target} (x_v^N - x_v^{ref})$, where $Q_{target} = w_{target}$.

*2) Comfort Cost:* The other objective of the AVs is to ensure ride comfort for the people inside the vehicle. Ride comfort is typically characterized by the jerk of the vehicle. Both sudden acceleration and sudden deceleration reduce ride comfort. Thus we penalize sudden changes in acceleration as $J_{jerk} = \Delta\mathbf{U}^T \mathbf{Q_{jerk}} \Delta\mathbf{U}$. Moreover, the acceleration is also penalized to restrict unnecessary acceleration or deceleration by the vehicle as $J_{acc} = \mathbf{U}^T \mathbf{Q_{acc}} \mathbf{U}$. The above quadratic costs are given by $\mathbf{Q_{jerk}} = diag(w_{jerk}, , w_{jerk})$, and $\mathbf{Q_{acc}} = diag(w_{acc}, , w_{acc})$.

*3) Speed Cost:* AVs are expected to follow the posted speed limit to maintain a smooth flow of traffic. Thus, we penalize the deviation from the reference speed as $J_{speed} = (\mathbf{V} - \mathbf{V_{ref}})^T \mathbf{Q_{speed}} (\mathbf{V} - \mathbf{V_{ref}})$, where, $\mathbf{Q_{speed}} = diag(w_{speed}, , w_{speed})$.

### E. Constraints

AV motion is constrained to follow the model discussed in equation (1). States and inputs are also constrained considering the physical limitations of the vehicle and to avoid potential collision with pedestrians. The different constraints developed are discussed below.

*1) AV Motion Model:* To ensure that the optimization problem calculates states and inputs that physically agree with the motion of the vehicle, the motion model mentioned in equation (1) is given as equality constraints.

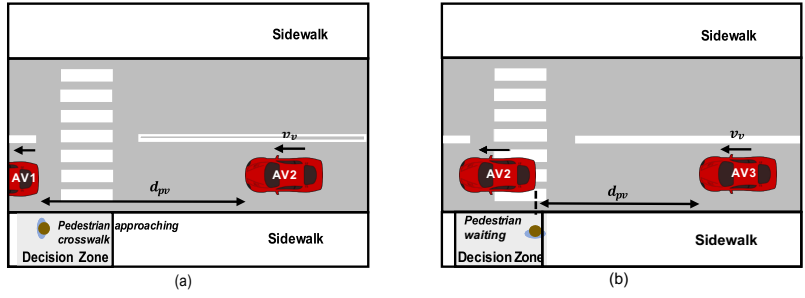

Fig. 3. Evaluation of gap acceptance: (a) pedestrian is approaching and close to the crosswalk and a gap starts and (b) pedestrian is waiting on the road and a gap starts. Gaps start when previous vehicle just crossed the pedestrian and when the pedestrian is in the decision zone. Pedestrians' decide to accept/reject the gaps only when they are in the Decision Zone and when a gap starts.

*2) State and Control Bounds:* Considering the physical limitations of the vehicle, we restrict the velocity, acceleration, and jerk of the vehicle represented as lower ($l_b$), and upper bounds ($u_b$) in equation (2).

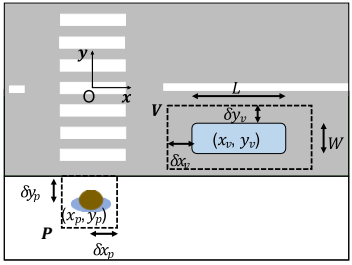

Fig. 4. Collision avoidance is incorporated by ensuring the sets **P** and **V**, representing the uncertain positions of AV and pedestrian respectively, do not intersect each other for the entire planning horizon.

*3) Collision Avoidance:* Pedestrian trajectory is predicted using the pedestrian crossing model discussed in Section III-C. Collision avoidance of the planned AV trajectory with the predicted pedestrian trajectory is ensured through inequality constraints in the optimization problem (refer equation (2)). We incorporate the uncertainty in the state estimation of pedestrians and vehicles as over-approximated rectangles, which is a conservative assumption that ensures safety. To avoid collision, the sets **P** and **V** (refer Fig. 4) should not intersect with each other at any time instant. This is expressed by the sets of inequality constraints in both *x* and *y* axes, represented by equations (4) and (5) respectively. A collision is avoided if at least one of the following four equations is satisfied at any given time.

$$x_p + \delta x_p \leq x_v - L/2 - \delta x_v$$
$$x_p - \delta x_p \geq x_v + L/2 + \delta x_v \quad (4)$$

$$y_p + \delta y_p \leq y_v - W/2 - \delta y_v$$
$$y_p - \delta y_p \geq y_v + W/2 - \delta x_v \quad (5)$$

We assume the AVs can accurately track the center line of the lane and thus neglect the lane boundary conditions in our formulation. The B-MPC controller is able to effectively combine pedestrian crossing behavior predictions for a long duration ($\approx$ 5 *s*) as constraints for collision avoidance and calculate inputs that optimize the AV's objectives.

## IV. SIMULATION

To evaluate the performance of the controller, we simulated a scenario where AVs are approaching an unsignalized mid-block crosswalk with a pedestrian possibly intending to cross the street (refer Fig. 5).

### A. Simulation setup

We simulate a midblock scenario with straight roads and assume the AVs follow the center line of the lane. We developed the simulation to be as realistic as possible by considering a stream of AVs approaching the crosswalk one after the other with varying speeds and time gaps. However, at any time, only one pedestrian will be in the simulation. AVs spawn with a random initial speed and a randomly varying time gap between their spawns. This ensures that there are both crossable and uncrossable gaps for the pedestrians. The simulation parameters are shown in Table II. The decision zone *D* is larger on the side of the approaching pedestrians (refer Table II). This ensures that approaching pedestrians have the opportunity to evaluate a new gap and decide to cross or wait whenever they are within *D*.

AVs assume that all pedestrians approaching the crosswalk have the intention of crossing the street until they walk past the crosswalk and out of the decision zone, *D* (refer Fig. 5). However, only a fraction of pedestrians ($\approx$ 80%) are randomly assigned the intention to cross the street. Pedestrians who have the intent to cross evaluate the gap when within the decision zone, whereas others just walk past the crosswalk at a constant velocity. Fig. 5 illustrates the AV – pedestrian interactions in the simulation. The gap of $AV_2$ will start immediately when $AV_1$ has crossed the pedestrian, at which point the pedestrian can decide to cross or wait.

### B. Baseline controller

The developed B-MPC is compared against a baseline rule-based controller. The baseline controller is a simple finite state machine (FSM) with four states: *Maintain Speed*, *Accelerate*, *Yield*, and *Hard Stop* (refer Fig. 6). The Boolean variable *InCW*, denotes the pedestrian's crossing activity.

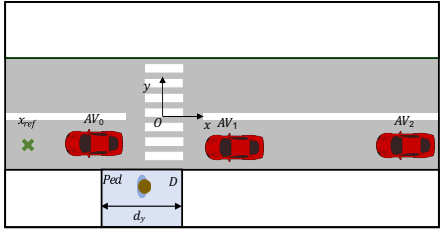

Fig. 5. Illustration of the AVs interacting with a pedestrian in the simulation. AVs' objective is to reach $x_{ref}$, given in Table II. Decision zone of pedestrians is represented by the set $D$ with length $d_y$.

*InCW* is 1 from the time the pedestrian started moving laterally to cross until they completely crossed the AV lane, and 0 otherwise. The variable $d_{comf}$ denotes the comfortable deceleration limit. The controller normally maintains the speed limit, $v_{ref}$. Whenever a pedestrian starts walking to cross the road, the controller always tries to stop, either by yielding or through hard stop. The deceleration is calculated as $a = -\frac{v_y^2}{2 dist_{stop}}$, where $v_v$ is the vehicle's current velocity and $dist_{stop}$ is the distance available to the AV before which it has to stop to avoid a collision. The stopped vehicle then accelerates back to its nominal speed once the pedestrian has crossed the AV's lane. The increments in the acceleration and deceleration at every time step are controlled by the comfortable jerk limits in the *Yield* state and by the hard jerk bounds in the *Hard Stop* state. Simulations are run with the same vehicle and pedestrian parameters shown in Table II.

TABLE II
PARAMETERS USED IN THE SIMULATION.

| Parameter | Value Range |
|---|---|
| Vehicle spawn speed [*m/s*] | 14 to 16 |
| Speed limit, $v_{ref}$ [*m/s*] | 16 |
| Spawn time gap between vehicles, $t_{spawn}$ [*s*] | 1 to 8 |
| Minimum time gap between vehicles to avoid collision, $t_{min}$ [*s*] | 2 |
| Hard speed bounds [*m/s*] | 0 to 50 |
| Comfortable acceleration limits, $d_{comf}, a_{comf}$ [*m/s²*] | -5 to 2 |
| Hard acceleration bounds [*m/s²*] | -10 to 10 |
| Comfortable jerk limits [*m/s³*] | -5 to 2 |
| Hard jerk bounds [*m/s³*] | -10 to 10 |
| Pedestrian decision zone length, $d_y$ [*m*] | -3 to 1 |
| Pedestrian speed [*m/s*] | 1 to 1.5 |
| AV destination, $P_{ref}[m, m]$ | (-120, -1.75) |
| Prediction horizon, $N$ [*s*] | 5 |
| Normalized cost function weights, | |
| $w_{target}$ | 0.004 |
| $w_{jerk}$ | 8 |
| $w_{acceleration}$ | 0.02 |
| $w_{speed}$ | 0.01 |

## V. RESULTS

The constrained quadratic control problem was solved using the standard quadratic program solver in MATLAB. The average run time of B-MPC with the prediction model was 24.7 *ms* with a standard deviation of 3.1 *ms*. The simulation was run for 500 pedestrians for both the B-MPC and the baseline cases as shown in Table III.

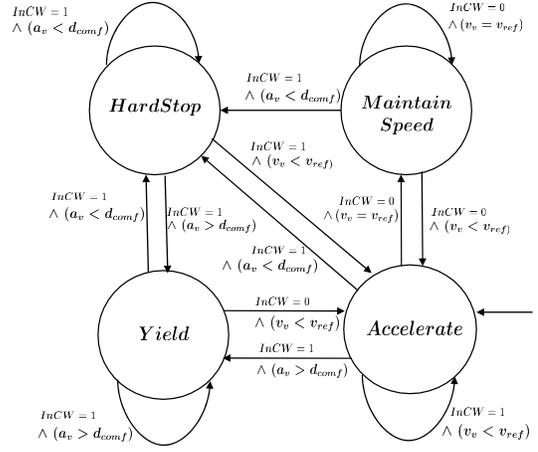

Fig. 6. Baseline Rule-based controller.

TABLE III
SIMULATION RUNS FOR B-MPC AND BASELINE CONTROLLERS.

| Parameter | B-MPC | Baseline |
|---|---|---|
| No. of Pedestrians | 500 | 500 |
| No. of AVs | 1434 | 1401 |
| No. of Crossings (Accepted gaps) | 411 | 405 |
| No. of pedestrians without crossing intent | 89 | 95 |

### A. B-MPC performance

Fig. 7 shows the B-MPC controller performance at various time instances for a nominal pedestrian interaction. The pedestrian in this case accepted a gap of 3.6 *s*. Initially, the AV travels at its preferred speed. The AV predicts the pedestrian is going to cross and reacts by starting to slow down at $t$ = 7.9 *s*. The AV starts to accelerate at $t$ = 11.0 *s* even before the pedestrian has crossed its lane. Finally, the AV goes past the pedestrian at $t$ = 13.3 *s*. The changes in speed can be seen through the changes in the spacing of the AV trajectory points (red points in Fig. 7). The long horizon prediction helps the AV in reacting early to the crossing pedestrian much before the pedestrian has actually started to walk or crossed the AV lane.

### B. Baseline comparison

We compared the B-MPC and the baseline controllers for varying time gaps and varying AV spawn speeds (refer Table II) for the cases when the pedestrians had the intent to cross. Fig. 8 compares the collision avoidance performance between the two controllers by evaluating the minimum distance to pedestrians ($d_{ped}$). It can be seen that the B-MPC has an overall higher minimum distance to pedestrian than the baseline case. For short gaps, the baseline controller sometimes is unable to avoid collisions (4 out of 500 cases). Whereas, the B-MPC controller avoids collisions for the range of gaps simulated. The B-MPC controller can thus handle a wider range of gaps and is applicable for a wider range of scenarios than the rule-based controller.

Fig. 9 compares other performance measures between the two controllers such as time to destination ($t_{des}$), average velocity ($v_m$), average acceleration ($a_m$), and average absolute jerk ($j_m$) during the interaction duration. Interaction duration was calculated between the time when the pedestrian started walking to cross, in the case of the baseline controller, or when the AV had predicted the start of pedestrian walking, in the case of the B-MPC, and the time when the pedestrian crossed and left the lane in which the AV travelled or when the AV had crossed the pedestrian, whichever occurred earlier. The overall average duration of interaction was higher for the B-MPC controller: $t = 6.08\ s$ for B-MPC and $t = 4.05\ s$ for the baseline controller. The B-MPC is able to reach its destination faster as it does not come to a complete stop unless necessary to avoid collisions, thereby improving the traffic flow. It can be seen that the B-MPC is more aggressive, efficient, and comfortable than the baseline as observed through the higher average velocity, lower average acceleration effort, and lower average jerk respectively.

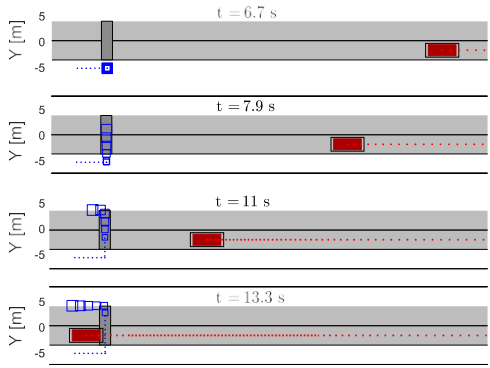

Fig. 7. A typical interaction between the AV and a pedestrian. The AV is represented by the red rectangle with the black rectangle indicating the position uncertainty. The red and the blue dots indicate the trajectories taken by the AV and the pedestrian respectively and the blue rectangles indicate the predictions of pedestrian trajectory by the AV. The AV starts slowing down at $t = 7.9\ s$ and starts accelerating at $t = 11\ s$, even before the pedestrian crosses the lane.

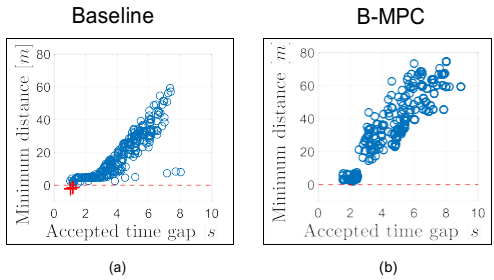

Fig. 8. Comparison of minimum distance to pedestrians between baseline and B-MPC controllers. The red 'plus' marks indicate the instances of collisions between the AV and the pedestrian. B-MPC controller is able to avoid collisions comfortably whereas collisions are inevitable at shorter gaps for the baseline controller.

### C. Non-crossing pedestrian interaction performance

We report the performance of both the B-MPC and the baseline controllers for the cases where the pedestrians did not intend to cross (refer Table IV). The B-MPC reacts to the approaching pedestrians within $D$ which can be seen from the slightly higher deceleration and reduced velocity. The baseline controller does not react at all since it never sees the pedestrian crossing laterally. Even still, the overall performance of the B-MPC is better, as seen by the lower deceleration, higher distance to pedestrian, and lower time to destination, than the baseline for our sample case where approximately 20% of pedestrians approaching the crosswalk did not intend to cross.

TABLE IV
PERFORMANCE METRICS FOR B-MPC AND BASELINE CONTROLLERS FOR PEDESTRIANS WITHOUT CROSSING INTENT AND ALL PEDESTRIANS.

| Parameter | B-MPC, No intent | Baseline, No intent | B-MPC, Overall | Baseline, Overall |
|---|---|---|---|---|
| $t_{int}$ [s] | 6.44 | 3.38 | 6.08 | 4.01 |
| $t_{des}$ [s] | 8.74 | 8.27 | 9.47 | 11.14 |
| $d_{ped}$ [m] | NA | NA | 33.7 | 20.1 |
| $v_m$ [m/s] | 13.9 | 14.8 | 13.3 | 9.3 |
| $a_m$ [m/s$^2$] | -0.10 | -0.04 | -0.51 | -1.90 |
| $j_m$ [m/s$^3$] | 0.04 | 0.04 | 0.11 | 0.18 |

## VI. CONCLUSIONS AND FUTURE WORK

In this paper we formulated the AV control problem for crosswalk interactions in an MPC based optimization framework. Unlike existing studies which assumed to have the future pedestrian information available [17], [24], we incorporated predictions of pedestrians crossing behavior from a previously developed pedestrian crossing model. The crossing model evaluated crossing behavior as a hybrid system with a gap acceptance model that required minimal information, namely pedestrian's position and velocity. By incorporating the crossing model, we were able to demonstrate the efficacy of the controllers in crosswalk situations with both waiting and approaching pedestrians. Existing studies, on the other hand, assumed that pedestrians were already present at the crosswalk to cross the street [24]. The implemented B-MPC is able to handle a variety of situations characterized by the wide range of pedestrian accepted gaps in the simulations without collisions. The framework is also able to plan safely and efficiently for long horizons with real-time performance.

The developed model and controller have certain limitations. The model assumed constant velocity within each discrete state which might be too simplistic. Also, the model assumed that all pedestrians have the intent to cross, which is not the case in the real-world. Further, the controller was only implemented in a simulated environment for a simple scenario and a simple AV motion model. Future work would focus on implementing the controller with the full model in more complex situations such as curved roads, intersections and with multiple pedestrians. Future work will also explore utilizing pedestrian data in real-world scenarios for model development and evaluation. Including AV intent communication to pedestrians to either yield or pass through the crosswalk in the MPC framework is yet another area to explore. Pedestrian crossing behavior, among other factors, depends on the available time gap and thus by safely increasing or decreasing the available time gap, the AVs can communicate their intent to the pedestrians.

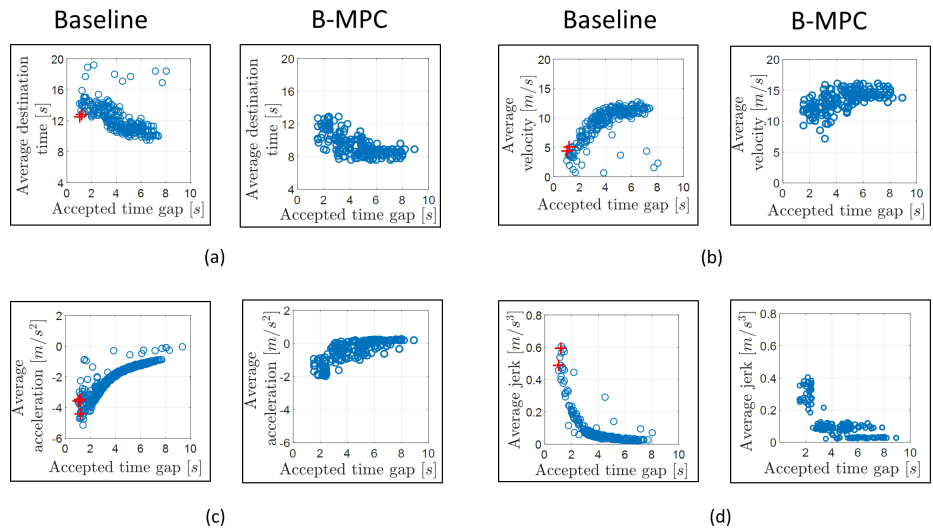

Fig. 9. Performance metrics comparison between baseline controller and B-MPC controller. The performance metrics compared are (a) time to destination, (b) average velocity, (c) average acceleration, and (d) average jerk during the interaction duration. The B-MPC controller is more efficient and comfortable as it results in less time to reach destination, less control effort, and less jerk than the baseline. The red 'plus' signs indicate the instances of collision.

## ACKNOWLEDGMENT

This research was supported in part by by the Automotive Research Center (ARC) at the University of Michigan, with funding from government contract DoD-DoA W56HZV14-2-0001, through the U.S. Army Combat Capabilities Development Command (CCDC) /Ground Vehicle Systems Center (GVSC) and in part by the National Science Foundation.

DISTRIBUTION A. Approved for public release; distribution unlimited. OPSEC# 3815

## REFERENCES

[1] L. Robert, "The future of pedestrian–automated vehicle interactions," *XRDS: Crossroads, ACM*, vol. 25, no. 3, 2019.
[2] A. Millard-Ball, "Pedestrians, Autonomous Vehicles, and Cities," *J. Planning Edu. Res.*, vol. 38, pp. 6–12, 2018.
[3] D. Sadigh, S. Sastry, S. A. Seshia, and A. D. Dragan, "Planning for autonomous cars that leverage effects on human actions." in *Robotics: Sci. Systems*, vol. 2. Ann Arbor, MI, USA, 2016.
[4] W. Schwarting, J. Alonso-Mora, and D. Rus, "Planning and decision-making for autonomous vehicles," *Annu. Review Control, Robot., Autonomous Systems*, 2018.
[5] C. G. Keller, T. Dang, H. Fritz, A. Joos, C. Rabe, and D. M. Gavrila, "Active pedestrian safety by automatic braking and evasive steering," *IEEE Trans. Intell. Transp. Syst.*, vol. 12, no. 4, pp. 1292–1304, 2011.
[6] M. A. Mousavi, Z. Heshmati, and B. Moshiri, "LTV-MPC based path planning of an autonomous vehicle via convex optimization," in *21st Iranian Conf. Elec. Engg.* IEEE, 2013, pp. 1–7.
[7] S. K. Jayaraman, D. Tilbury, X. J. Yang, A. K. Pradhan, and L. P. Robert Jr., "Analysis and prediction of pedestrian crosswalk behavior during automated vehicle interactions," in *IEEE Int. Conf. Robot. Autom.*, 2020.
[8] H. Bai, S. Cai, N. Ye, D. Hsu, and W. S. Lee, "Intention-aware online POMDP planning for autonomous driving in a crowd," in *IEEE Int. Conf. Robot. Autom.* IEEE, 2015, pp. 454–460.
[9] I. Batkovic, M. Zanon, M. Ali, and P. Falcone, "Real-time constrained trajectory planning and vehicle control for proactive autonomous driving with road users," *arXiv preprint arXiv:1903.07743*, 2019.
[10] M. Goldhammer, S. Köhler, K. Doll, and B. Sick, "Camera based pedestrian path prediction by means of polynomial least-squares approximation and multilayer perceptron neural networks," in *SAI Intell. Systems Conf.* IEEE, 2015, pp. 390–399.
[11] R. Quintero Minguez, I. Parra Alonso, D. Fernandez-Llorca, and M. A. Sotelo, "Pedestrian Path, Pose, and Intention Prediction Through Gaussian Process Dynamical Models and Pedestrian Activity Recognition," *IEEE Trans. Intell. Transp. Syst.*, vol. 20, no. 5, pp. 1803–1814, 2018.
[12] B. Volz, K. Behrendt, H. Mielenz, I. Gilitschenski, R. Siegwart, and J. Nieto, "A data-driven approach for pedestrian intention estimation," in *IEEE Int. Conf. Intell. Transp. Syst.*, 2016, pp. 2607–2612.
[13] A. E. Sallab, M. Abdou, E. Perot, and S. Yogamani, "Deep reinforcement learning framework for autonomous driving," *Electronic Imaging*, vol. 2017, no. 19, pp. 70–76, 2017.
[14] Y. Chen and J. Wang, "Trajectory tracking control for autonomous vehicles in different cut-in scenarios," in *American Control Conf.* IEEE, 2019, pp. 4878–4883.
[15] Y. Gao, A. Gray, A. Carvalho, H. E. Tseng, and F. Borrelli, "Robust nonlinear predictive control for semiautonomous ground vehicles," in *American Control Conf.* IEEE, 2014, pp. 4913–4918.
[16] A. Gray, Y. Gao, T. Lin, J. K. Hedrick, H. E. Tseng, and F. Borrelli, "Predictive control for agile semi-autonomous ground vehicles using motion primitives," in *American Control Conf.* IEEE, 2012, pp. 4239–4244.
[17] M. Werling and D. Liccardo, "Automatic collision avoidance using model-predictive online optimization," in *IEEE Conf. Decision Control.* IEEE, 2012, pp. 6309–6314.
[18] J. Kong, M. Pfeiffer, G. Schildbach, and F. Borrelli, "Kinematic and dynamic vehicle models for autonomous driving control design," in *IEEE Intell. Veh. Symp.* IEEE, 2015, pp. 1094–1099.
[19] J. F. Kooij, F. Flohr, E. A. Pool, and D. M. Gavrila, "Context-Based Path Prediction for Targets with Switching Dynamics," *Int. J. Comput. Vision*, vol. 127, no. 3, pp. 239–262, 2019.
[20] G. Yannis, E. Papadimitriou, and A. Theofilatos, "Pedestrian gap acceptance for mid-block street crossing," *Transp. Planning Technol.*, vol. 36, no. 5, pp. 450–462, 2013.
[21] John C. Platt, "Probabilistic Outputs for Support Vector Machines and Comparisons to Regularized Likelihood Methods," *Advances Large Margin Classifiers*, vol. 10, no. 3, pp. 61–74, 1999.
[22] A. Rasouli, I. Kotseruba, and J. K. Tsotsos, "Understanding Pedestrian Behavior in Complex Traffic Scenes," *IEEE Trans. Intell. Veh.*, vol. 3, no. 1, pp. 61–70, 2018.
[23] S. K. Jayaraman, C. Creech, D. M. Tilbury, X. J. Yang, A. K. Pradhan, K. M. Tsui, and L. P. Robert, "Pedestrian trust in automated vehicles: Role of traffic signal and av driving behavior," *Frontiers in Robotics and AI*, vol. 6, p. 117, 2019.
[24] N. R. Kapania, V. Govindarajan, F. Borrelli, and J. C. Gerdes, "A Hybrid Control Design for Autonomous Vehicles at Uncontrolled Intersections," *arXiv preprint arXiv:1902.00597*, 2019.